\documentclass[conference,onecolumn]{IEEEtran}

\usepackage{graphicx}

\usepackage{tikz}
\usepackage{comment}
\usepackage{amsmath,amssymb} 
\usepackage{color}
\usepackage{xcolor}
\usepackage{booktabs}
\usepackage{multirow}
\usepackage{pgfplots}
\usepackage{bbm}
\usepackage{enumitem}

\newtheorem{definition}{Definition}

\usepackage{pifont}
\newcommand{\xmark}{\ding{55}}
\newcommand{\cmark}{\ding{51}}

\usetikzlibrary{pgfplots.groupplots}
\usetikzlibrary{patterns}
\pgfplotsset{
    every non boxed x axis/.style={} 
}

\definecolor{bblue}{HTML}{4F81BD}
\definecolor{rred}{HTML}{C0504D}
\definecolor{ggreen}{HTML}{9BBB59}
\definecolor{ppurple}{HTML}{9F4C7C}

\usepackage[accsupp]{axessibility}  

\begin{document}
\title{PNM: Pixel Null Model for General Image Segmentation}

\author{
Han Zhang$^1$, Zihao Zhang$^1$, Wenhao Zheng$^2$, and Wei Xu$^1$\\
$^1$ Institute for Interdisciplinary Information Sciences, Tsinghua University, $^2$ ByteDance Group\\
}

%
%


\maketitle
\begin{abstract}

A major challenge in image segmentation is classifying object boundaries.
Recent efforts propose to refine the segmentation result with boundary masks.
However, models are still prone to misclassifying boundary pixels even when they correctly capture the object contours.
In such cases, even a perfect boundary map is unhelpful for segmentation refinement.
In this paper, we argue that assigning proper prior weights to error-prone pixels such as object boundaries can significantly improve the segmentation quality. 
Specifically, we present the \textit{pixel null model} (PNM), a prior model that weights each pixel according to its probability of being correctly classified by a random segmenter.
Empirical analysis shows that PNM captures the misclassification distribution of different state-of-the-art (SOTA) segmenters.
Extensive experiments on semantic, instance, and panoptic segmentation tasks over three datasets (Cityscapes, ADE20K, MS COCO) confirm that PNM consistently improves the segmentation quality of most SOTA methods (including the vision transformers) and outperforms boundary-based methods by a large margin.
We also observe that the widely-used mean IoU (mIoU) metric is insensitive to boundaries of different sharpness.
As a byproduct, we propose a new metric, \textit{PNM IoU}, which perceives the boundary sharpness and better reflects the model segmentation performance in error-prone regions.
\end{abstract}

\section{Introduction}
\label{sec:intro}

Image segmentation is a fundamental computer vision task that has a wide range of applications in autonomous driving~\cite{DBLP:journals/tits/FengHRHGTWD21,DBLP:conf/cvpr/YangYZLY18}, medical image analysis~\cite{DBLP:conf/3dim/MilletariNA16,DBLP:conf/miccai/ZhouSTL18}, virtual reality~\cite{DBLP:conf/cvpr/0012KGFK19} etc.
Based on the pioneering fully convolutional network (FCN)~\cite{fcn}, current image segmentation models are able to classify every input pixel, which enables pixel-level recognition of the objects and overall semantic understanding of the image.

\begin{figure}[!tb]
\centering
\includegraphics[width=\linewidth]{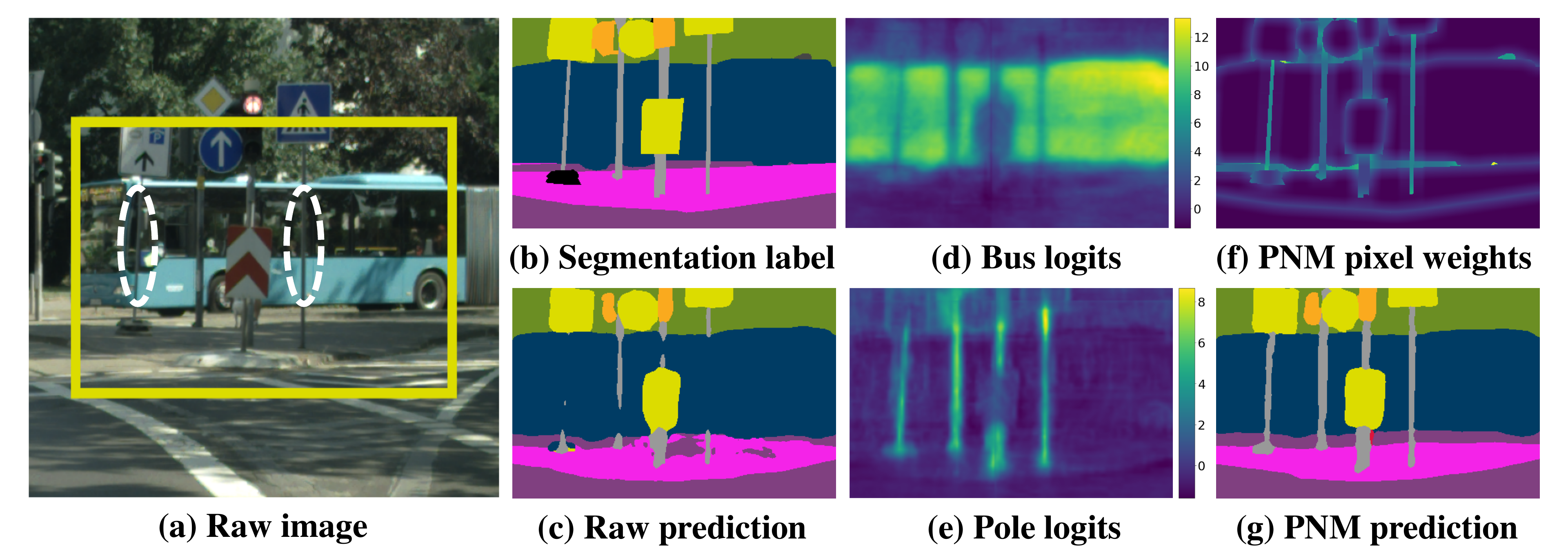}
\caption{
An example showing a model can misclassify boundary pixels even if it correctly captures the object contour.
(a) shows the raw image from the Cityscapes~\cite{cityscapes} \textit{val} benchmark and (b) shows the corresponding ground truth segmentation mask.
(c) shows the prediction by DeepLabV3+~\cite{deeplabv3_plus} that misses much of the poles.
However, if we look at the class logits (i.e., the input of the softmax layer before calculating losses) in (d) and (e), we see that DeepLabV3+ correctly captures object contour.
In fact, in (c) and (d), we can see that even the local minima of bus logits are greater than the local maxima of pole logits, which means the imbalanced logit scale of adjacent classes causes the misclassification.
This fact suggests that we should assign larger weights to the error-prone pole pixels (i.e., boundary pixels).
(f) shows the pixel weights our pixel null model (PNM) assigns, where brighter pixels have larger weights.
(g) shows that DeepLabV3+ with PNM can perfectly segment the poles.
Note that different pole instances have different weights (brightness), and we will analyze them in Section~\ref{sec:pnm_vis}.
Best viewed on screen.
}
\label{fig:motivation}
\end{figure}

Despite great progress, a big gap still exists between the segmentation quality of current models and humans.
Recent efforts find that the major performance loss of the state-of-the-art (SOTA) segmenters lies in the misclassification of boundary pixels~\cite{segfix}.
To solve this problem, a line of research~\cite{DBLP:conf/iccv/DingJLM019,DBLP:conf/iccv/TakikawaAJF19,segfix,DBLP:conf/cvpr/ZhenWZLSSFQ20,csel} proposes to refine the segmentation results with a boundary mask calculated on the segmentation labels.
However, such methods may fail to refine the segmentation results of SOTA segmenters even with a perfect boundary mask. ~\figurename~\ref{fig:motivation} shows an example.  \figurename~\ref{fig:motivation}(c), (d), (e) shows that even if the model captures the object contours, the unbalanced logit values of the adjacent classes still cause misclassification, rendering the boundary-aware components or branches of the network less significant in boosting prediction quality. 

In this paper, we argue that assigning proper weights to error-prone pixels such as object boundaries for training can significantly improve the segmentation quality.
We propose the \textit{pixel null model (PNM)}, a prior model that weights each pixel according to its probability of being correctly classified by a random segmenter.
This random segmenter permutes the ground truth labels within a local image patch and outputs the permuted labels as the predicted segmentation mask.
PNM then calculates the expected accuracy of this random segmenter as a prior distribution of the correct classification probability for general models.

We name the model \textit{pixel null model} as it shares a similar idea with the well-known \textit{null model} in community detection problem on graphs.
Newman and Girvan~\cite{Newman2004FindingAE,Newman2006ModularityAC} propose to evaluate the quality of any community assignment by comparing the edge density within the communities with the expected edge density of a \textit{null model} that randomly redistributes the edges.  
During the permutation, the degree of each node stays the same, which is analogous to the local pixel label permutation in PNM.

When applying to model training, we show with empirical analysis that SOTA segmentation networks such as Segformer~\cite{segformer}, OCRNet~\cite{ocrnet}, and DeepLabV3~\cite{deeplabv3} are more likely to misclassify the pixels with higher PNM weights.
Based on pixel weights derived from PNM, segmentation networks can focus on error-prone pixels for a better allocation of their model capacity (\figurename~\ref{fig:motivation}(f), (g)).

We perform extensive experiments on three main image segmentation tasks (i.e., semantic~\cite{fcn}, instance~\cite{mask-rcnn}, and panoptic~\cite{panoptic_seg} segmentation) on Cityscapes~\cite{cityscapes}, ADE20k~\cite{ade20k} and MS COCO~\cite{mscoco} datasets.
PNM brings significant improvements to SOTA models, including the recently proposed vision transformers~\cite{vit,segformer} on all three tasks and datasets, and outperforms the boundary-based methods by a large margin.

Moreover, we show that a PNM-based metric \textit{PNM IoU} is also helpful in model evaluation, as it overcomes the sharp boundary identifiability issue of IoU-based metrics (Section~\ref{sec:pnm_iou}) and more accurately reflects the segmentation performance.
We summarize our contributions as follows.
\begin{itemize}
	\item We identified an important cause of boundary misclassification, where even a perfect boundary mask is unhelpful for segmentation refinement;
    \item We propose PNM, a pixel-level prior model of correct segmentation probability that captures the misclassification distribution of SOTA networks;
    \item As a byproduct, we propose the PNM IoU metric, which overcomes the sharp boundary identifiability issue of conventional IoU-based metrics;
    \item The significant improvement over SOTA segmentation networks on three image segmentation tasks and three benchmark datasets demonstrate the general effectiveness of PNM.
\end{itemize}

\section{Related Work}
\label{sec:related_work}

\subsection{Image Segmentation}
\label{sec:image_seg}

Image segmentation aims to classify each pixel into different semantic categories.  There are three main task types:
1) Instance segmentation~\cite{mask-rcnn,panet} mainly focuses on things (i.e., countable objects) in the image and does not predict segmentation masks for stuff (i.e., uncountable objects such as the sky).  It is a natural extension of the object detection task~\cite{faster-rcnn}.
For each pixel, it predicts a binary mask that indicates whether the pixel belongs to the detected object.
2) Semantic segmentation~\cite{fcn,deeplab} requires classifying every pixel of the image into multiple classes, including both things and stuff.
3) Panoptic segmentation~\cite{panoptic_seg} combines the two tasks, not only recognizing both things and stuff, but also distinguishing instances of the same thing class at the same time.
Boundary accuracy is essential for all three tasks.  

FCN~\cite{fcn} is the \emph{de facto} workflow standard for current image segmentation, i.e. we first densely predict the segmentation mask and then calculate the classification loss independently for each pixel.
Many models provide significant improvements based on FCN, e.g., enlarging the receptive field through dilated convolution~\cite{deeplab,deeplabv3,deeplabv3_plus}, fusing information from multiple scales~\cite{deeplabv3_plus,hrnet,upernet,pspnet}, and introducing attention mechanism into segmentation heads~\cite{danet,psanet,ccnet,ocrnet,nlnet} or backbones~\cite{vit,swin,segformer,setr}.
One drawback of these methods is that they treat all pixels indiscriminately throughout the entire learning process (e.g., the shared convolutional kernels and the same weight for calculating losses).
This leads to sub-optimal models as a small subset of pixels contributes much more significantly to image semantics, and pixels on boundaries are examples.
We use PNM to increase weights to these pixels and thus allowing the model to focus on error-prone regions so as to make better use of the fitting ability of the models.

\subsection{Boundary-aware Methods}
\label{sec:boundary_methods}

Recent empirical analysis reveals that segmentation models are more likely to misclassify the pixels near object boundaries~\cite{segfix}.
Many researchers propose to use the boundary information of objects to refine the segmentation results~\cite{DBLP:conf/iccv/DingJLM019,DBLP:conf/iccv/TakikawaAJF19,segfix,DBLP:conf/cvpr/ZhenWZLSSFQ20}.
There are two types of boundary-aware methods.  
The first type also predict whether the pixel lies on object boundaries in addition to the segmentation mask~\cite{DBLP:conf/cvpr/BorseWZP21,DBLP:conf/nips/ShenJLWL20,DBLP:conf/iccv/TakikawaAJF19,segfix,DBLP:conf/cvpr/ZhenWZLSSFQ20}.
The second type encodes boundary information into the model and limits the exchange of information across boundaries~\cite{DBLP:conf/iccv/DingJLM019,csel}.

Unfortunately, it is difficult for segmenters of either type to visually recognize the ``boundary pixels'' due to three reasons:
1) the labels of the pixels are sensitive to the choice of the boundary width;
2) there is a large variety of local texture features of boundary pixels, as it depends on the combination of classes on both sides of the boundary;
3) only a small subset of pixels lie on boundaries, and thus the training process is highly unbalanced. 
In addition, using a model to predict boundary pixels introduces nontrivial additional computational overhead, both for training and inference.
Even worse, our example in \figurename~\ref{fig:motivation} in Section~\ref{sec:intro} shows that even with perfect boundary maps, we may still fail to segment correctly due to the unbalanced logit value issue.

PNM takes a completely different approach by not explicitly defining the semantic ``boundaries'', but only assigning larger weights to ``error-prone'' pixels in terms of recognizability of a random model.  
PNM is essentially the pixel weights for loss calculation, and thus easy to plug into most SOTA models, with only a negligible computational overhead for training and no cost for inference.

\subsection{Edge Null Model for Community Detection on Graphs}
\label{sec:enm}

At a high level, the idea of PNM is similar to the modularity method~\cite{Newman2006ModularityAC} in community detection problems, in that both use a random model to assist the learning of the main model.
The goal of community detection is to cluster the vertices of a graph into compactly connected \textit{communities}.
A major challenge is how to quantitatively evaluate the \emph{compactness} (a.k.a. \emph{modularity}) of a given community assignment~\cite{DBLP:journals/corr/abs-0906-0612}, especially when the edges have the same weights.


Newman and Girvan introduce the \textit{edge null model (ENM)} to distinguish the contribution of different edges to the community structure~\cite{Newman2004FindingAE}.
For any pair of vertices, ENM is defined as the probability that an edge exists between these two vertices after randomly redistributing the edges of the graph, while keeping the degree of each vertex unchanged during the random permutation\footnote{We provide an illustration of ENM in the supplements.}.
Edges with a lower probability of presence in ENM are considered stronger, as their presence brings richer information about the affinity of the connected vertices.
Intuitively, two recruiters with over 20K connections each on LinkedIn know each other, but the fact offers a much weaker indication that they belong to the same community, compared to the fact of two self-isolated nerds knowing each other.  That is,  the edge between these two well-connected recruiters contributes little to the community structure, as the amount of information brought by the edge is limited.  
Comparing the edge density of the communities with the expected edge density of ENM, ~\cite{Newman2006ModularityAC} proposes an effective way to evaluate the community compactness.  
In our image segmentation tasks, the interior pixels of large objects contribute to the segmentation quality much less than boundary pixels.

\section{Pixel Null Model}
\label{sec:pnm}

\subsection{Definition}
\label{sec:pnm_definition}

\begin{figure}[!tb]
\centering
\includegraphics[width=\linewidth]{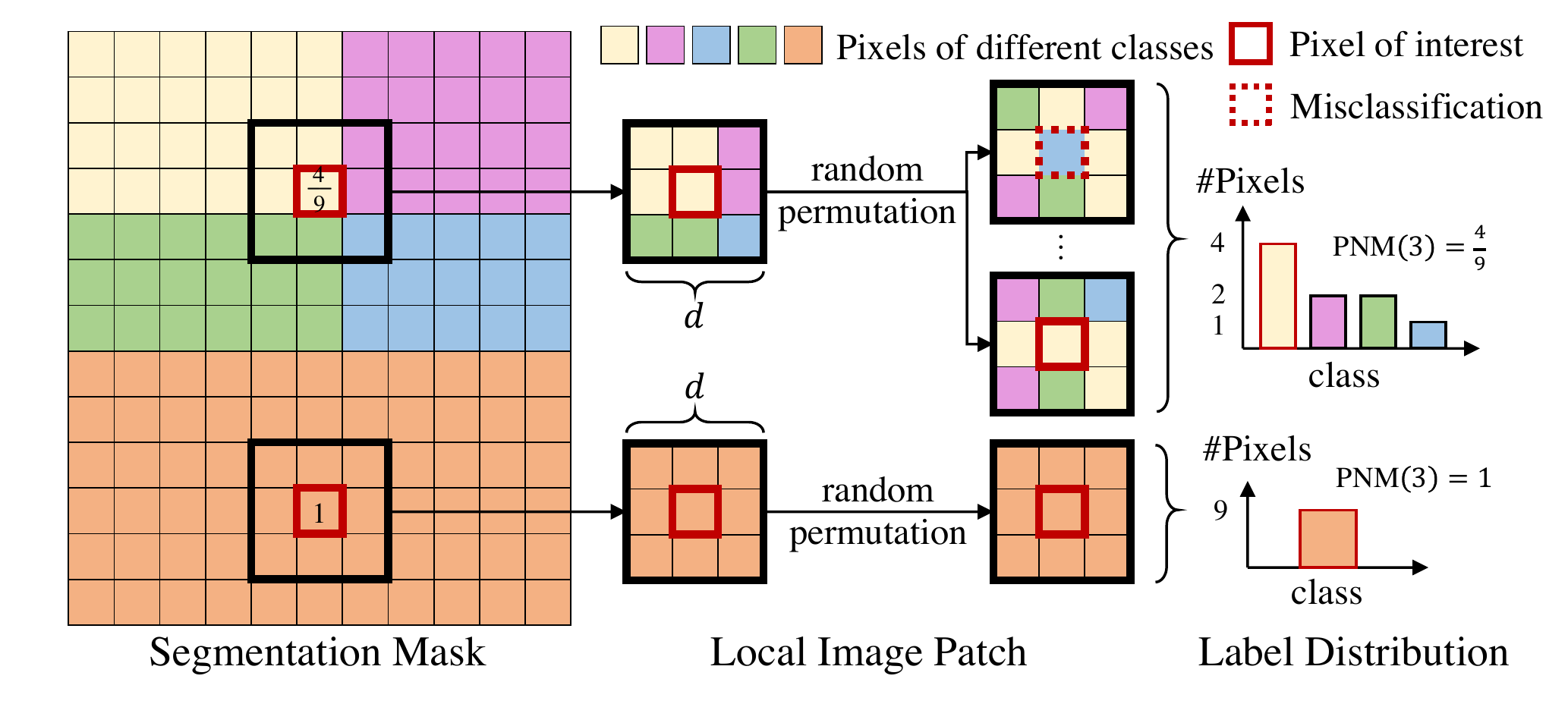}
\caption{
Illustration of PNM with locality $d=3$.
The PNM of each pixel is the probability that the pixel has the same semantic label after the local permutation of the segmentation labels within the local image patch.
During the permutation, PNM preserves the local pixel label distribution.
Best viewed on screen.
}
\label{fig:pnm}
\end{figure}

The success of convolutional neural networks shows that the local pixel relationships are vital to the semantic understanding of images.
Even for the recent vision transformers, the first layer of the model is essentially a strided convolutional layer~\cite{vit}. 
Inspired by ENM, we want to distinguish the contribution of each pixel to image semantics.
We propose the \textit{pixel null model} (PNM), a prior model that weights each pixel according to its probability of being correctly classified by an auxiliary random segmenter. \figurename~\ref{fig:pnm} illustrates the idea of PNM.

\begin{definition}[Pixel null model]
    \label{def:pnm}
    For pixel $i$ and the local square image patch $\mathcal{D}$ with side
    length d centered on i, the pixel null model of $i$ with locality scale $d$ is
    the probability that the semantic label $t_i$ of $i$ stays the same after
    a random permutation of the semantic labels within $\mathcal{D}$
    \begin{equation}
        \text{PNM}_i(d) \triangleq \mathbb{P}(t_i=t'_i),
    \end{equation}
    where $t'_i$ is the label of $i$ after permutation.
\end{definition}

In practice, according to Definition~\ref{def:pnm}, we can easily calculate PNM by counting the number of pixels with the same label as the pixel of interest within the local image patch
\begin{equation}
    \label{eq:pnm}
    \text{PNM}_i(d) = \frac{\sum_{j\in\mathcal{D}}\delta(t_i, t_j)}{d^2},~\delta(x, y) = \text{$1$ if $x = y$, otherwise $0$}.
\end{equation}
PNM characterizes a prior probability distribution of correct image segmentation.
To apply this prior distribution to training or evaluation, we can assign larger weights to pixels more prone to misclassification according to PNM.
In this paper, we consider three types of transformations from PNM to pixel weights and analyze the effect of these transformations in Section~\ref{sec:ablation_exp}.
\begin{align}
    \rho^{(1)}_i(d) &= -\log(\text{PNM}_i(d)) + 1, \label{eq:pnm_weight_1} \\
    \rho^{(2)}_i(d) &= 2 - \text{PNM}_i(d), \label{eq:pnm_weight_2}\\
    \rho^{(3)}_i(d) &= \frac{1}{\text{PNM}_i(d)}, \label{eq:pnm_weight_3}
\end{align}
where  $\rho^{(1)}_i(d), \rho^{(3)}_i(d)\in[1, +\infty)$ and $\rho^{(2)}_i(d)\in[1, 2]$.

\subsection{Visualization and Discriminative Properties}
\label{sec:pnm_vis}

\begin{figure}[!tb]
\centering
\includegraphics[width=\linewidth]{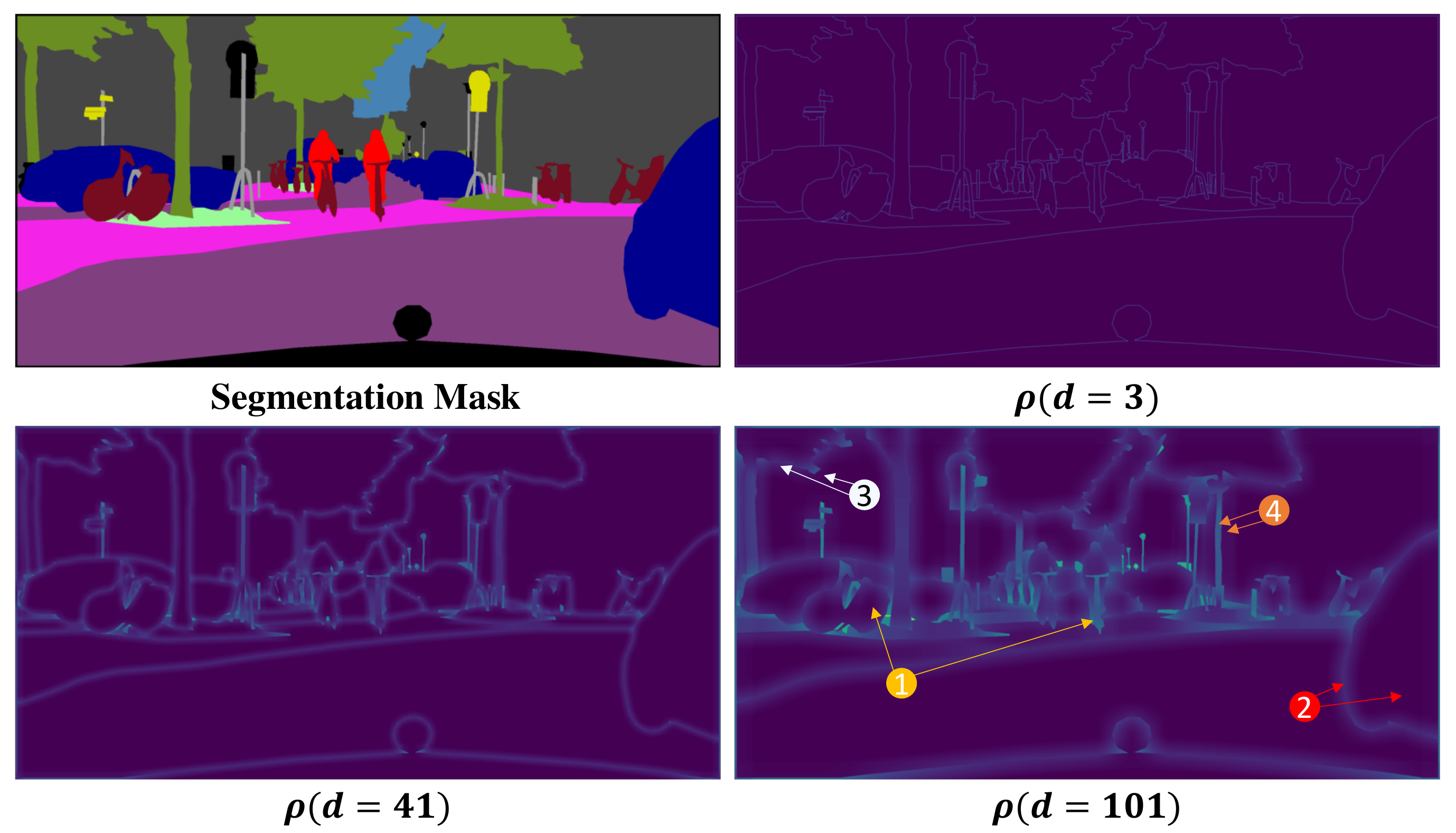}
\caption{
Illustration of PNM with different locality scales.
We use Eq.~\eqref{eq:pnm_weight_1} to transform PNM into pixel weights $\rho^{(1)}$.
Brighter pixels indicate larger weights.
The circled numbers mark the discriminative properties of PNM (Section~\ref{sec:pnm_vis}).
Best viewed on screen.
}
\label{fig:pnm_scale}
\end{figure}

The locality scale $d$ of PNM determines the scale of the local image patches for calculation.
\figurename~\ref{fig:pnm_scale} visualizes PNM under different locality scale $d$.
Obviously, when $d=1$, PNM treats all pixels equally, as the random permutation will not change the pixel label.
When $d=3$, the local image patches are smaller than most objects.
Therefore, the permutation will not affect the labels of interior pixels, and we can see that PNM outlines the thin contours of the objects.
As $d$ increases, due to the enlarged size of the local image patches, more object pixels appear in the local image patches, and the contours of the object begin to blur.
This indicates that the PNM perceives pixels within a certain distance near the boundaries.
Meanwhile, small and sharp objects gradually become brighter, showing that PNM gradually captures the size and shape information.

Although the definition of PNM does not explicitly involve object boundaries, PNM successfully captures the boundary information through the random segmenter.
However, unlike the existing boundary-based methods, PNM gets rid of the ambiguity brought by the vague definition of the boundaries (e.g., the width of boundaries) and, at the same time, distinguishes the pixels in a more detailed manner.
Specifically, PNM has the following four discriminative properties.  
\begin{enumerate}[label=\textbf{P\arabic{*}.}, leftmargin=2em]
    \item multiple instances of the same class (larger for smaller instances);
    \item interior vs. boundary pixels in an instance (larger for boundary pixels);
    \item pixels along a single boundary (larger for parts with sharp corners);
    \item classes on different sides of the boundary (larger for smaller classes).
\end{enumerate}

\begin{figure}[!tb]
\centering
\includegraphics[width=\linewidth]{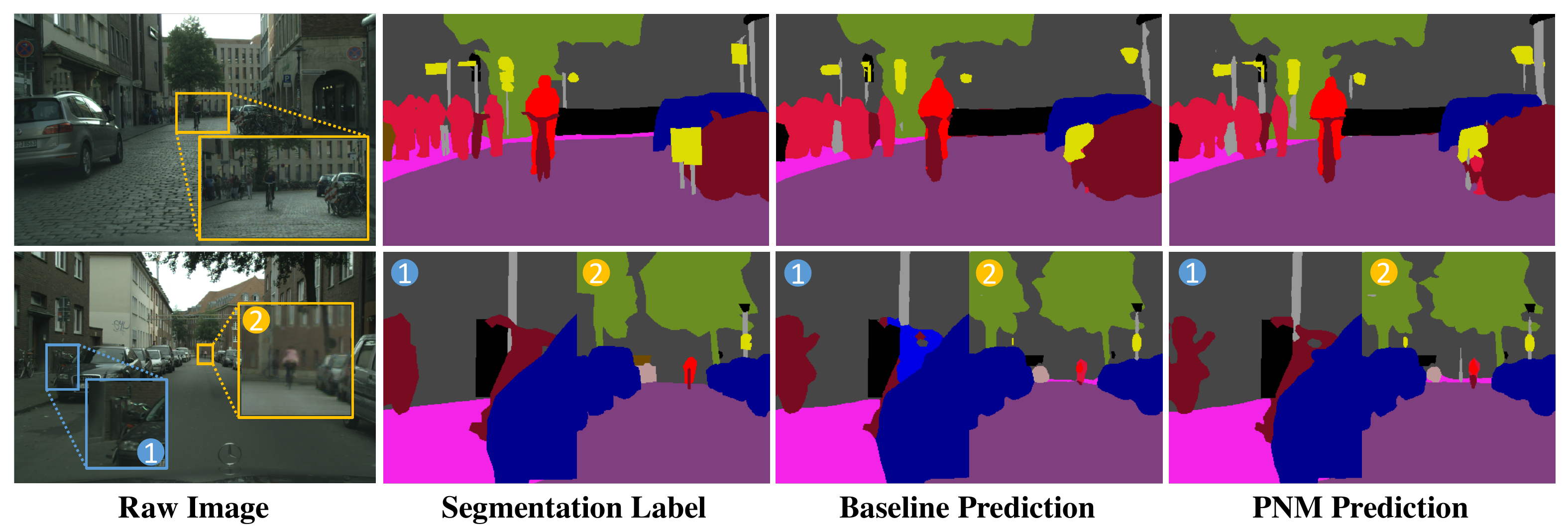}
\caption{
Qualitative examples of PNM on Cityscapes \textit{val}.
Best viewed on screen.
}
\label{fig:quality}
\end{figure}

The circled numbers in \figurename~\ref{fig:pnm_scale} highlight the four main discriminative properties of PNM.
(P1) Note that two bicycles are different w.r.t. PNM, with higher weight (brighter) for the more distant and sharp instance.
It helps the segmenter to detect small objects.  
(P2) The boundary of the car is brighter than the interior.
(P3) The boundary pixels of vegetation have different brightness, as the boundary contour has many sharp corners and the corners are brighter. 
(P4) The trunk of the tree is much brighter than the background building, as the trunk is the smaller class, which is particularly important as it helps to solve the class imbalance problem in Section~\ref{sec:intro}.
\figurename~\ref{fig:quality} gives examples that show PNM improves the segmentation quality of the baseline segmenter on error-prone pixels.

\section{PNM for Model Training and Evaluation}

In this section, we first show that PNM-based pixel weights are useful in model training as PNM captures the error distribution of SOTA segmentation networks.
Then we introduce a new metric for evaluation, \textit{PNM IoU} based on PNM, which overcomes the sharp boundary identifiability issue of IoU-based metrics.

\subsection{PNM for Model Training}

PNM describes the probability that a random segmenter predicts the correct segmentation mask when the segmenter is aware of the local semantic class distribution.
Therefore, PNM equivalently models the pixel-level misclassification probability of objects with different shapes and sizes.

\begin{figure}[!tb]
\centering
\begin{tikzpicture}
\begin{axis}[
    width=0.33\textwidth, height=0.28\textwidth,
    ymin=0, ymax=87, xmin=1, xmax=5,
    ylabel near ticks, ylabel shift={-5pt},
    xlabel near ticks, xlabel shift={-5pt},
    xlabel={Pixel Weight},
    ylabel={Error Rate (\%)},
    title={DeepLabV3},
    title style={yshift=-1.5ex},
    minor y tick num = 3,
    area style,
    ]
\addplot+[ybar interval,mark=no] table [x, y, col sep=comma] {data/error_rate_deeplabv3.csv};
\end{axis}
\end{tikzpicture}
\begin{tikzpicture}
\centering
\begin{axis}[
    width=0.33\textwidth, height=0.28\textwidth,
    ymin=0, ymax=87, xmin=1, xmax=5,
    ylabel near ticks, ylabel shift={-5pt},
    xlabel near ticks, xlabel shift={-5pt},
    xlabel={Pixel Weight},
    ylabel={Error Rate (\%)},
    title={OCRNet},
    title style={yshift=-1.5ex},
    minor y tick num = 3,
    area style,
    ]
\addplot+[ybar interval,mark=no] table [x, y, col sep=comma] {data/error_rate_ocr.csv};
\end{axis}
\end{tikzpicture}
\begin{tikzpicture}
\centering
\begin{axis}[
    width=0.33\textwidth, height=0.28\textwidth,
    xmin=1, xmax=5, ymin=0, ymax=87,
    ylabel near ticks, ylabel shift={-5pt},
    xlabel near ticks, xlabel shift={-5pt},
    xlabel={Pixel Weight},
    ylabel={Error Rate (\%)},
    title style={yshift=-1.5ex},
    title={Segformer},
    minor y tick num = 3,
    area style,
    ]
\addplot+[ybar interval,mark=no] table [x, y, col sep=comma] {data/error_rate_segformer.csv};
\end{axis}
\end{tikzpicture}
\caption{
Relationship between PNM-derived pixel importance and misclassification rate of DeepLabV3~\cite{deeplabv3}, OCRNet~\cite{ocrnet}, and Segformer~\cite{segformer} on Cityscapes \textit{val}.
The pixels with larger weight are more likely to be misclassified by a neural network.
}
\label{fig:pnm_stats}
\end{figure}
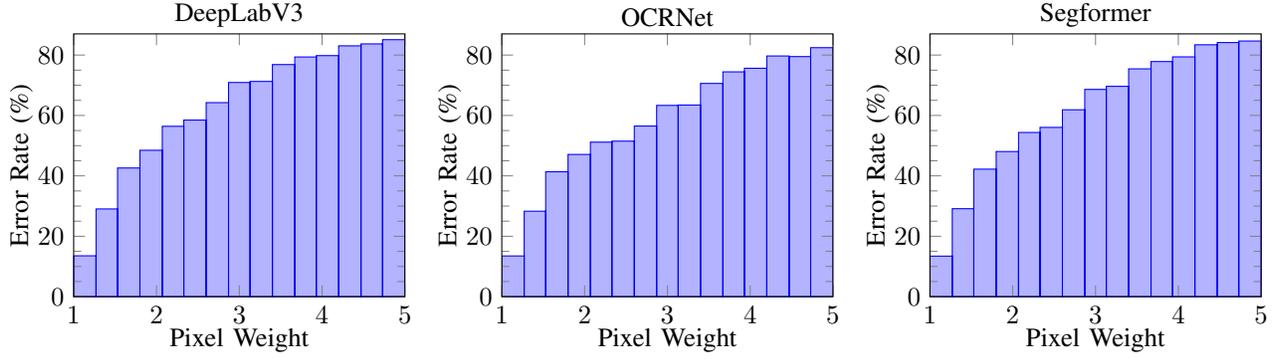
Interestingly, we find that even SOTA models make lots of mistakes on pixels with high PNM weights, as \figurename~\ref{fig:pnm_stats} shows.
Specifically, we calculate the pixel weights on Cityscapes \textit{val} according to Eq.~\eqref{eq:pnm_weight_1}, and then count the error rate of three SOTA models, DeepLabV3~\cite{deeplab},
OCRNet~\cite{ocrnet} and Segformer~\cite{segformer}, in different weight intervals.
We find that all three strong segmenters tend to misclassify the pixels with higher PNM weights, indicating that PNM-based pixel weights are high-quality pixel-level prior for SOTA segmenters.
Incorporating the weight into loss calculation allows the segmenters to stress more on error-prone pixels.
This observation is consistent with the previous finding in ~\cite{segfix} that object boundaries are error-prone for segmentation networks, except that PNM distinguishes pixels at a finer granularity than just interior-boundary pixels.

In practice, suppose the per pixel loss function is $\mathcal{L}: \mathbb{R}\times\{1,\dots,K\}\rightarrow\mathbb{R}$, the total number of classes and pixels are $K$ and $N$, $y_i$ and $t_i$ are the prediction and label for pixel $i$.
We calculate segmentation loss based on PNM as
\begin{equation}
    \text{PNM Loss}(d) = \frac{1}{N}\sum_{i=1}^{N}\rho_i(d)\cdot\mathcal{L}(y_i, t_i),
\end{equation}
which is essentially a pixel-wise weighted version of $\mathcal{L}$.
Note that PNM only slightly increase training time or space complexity, as we need to calculate the pixel weights once in preprocessing.
PNM has no effects on inference cost.

\subsection{PNM for Model Evaluation}
\label{sec:pnm_iou}

~\cite{DBLP:conf/cvpr/ChengGDBK21} finds that current metrics encourage models to better segment the interior pixels and may tolerate errors on boundaries, as the number of boundary pixels grows linearly with object size while that of interior pixels grows quadratically.

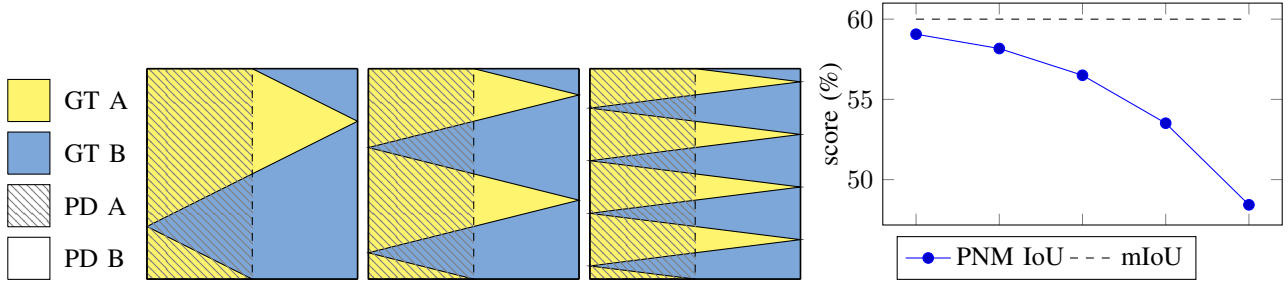
\begin{figure}[!tb]
\centering
\begin{tikzpicture}[scale=0.7]
  \filldraw[fill=white!30!yellow] (0,3) -- (0,3.8) -- (0.8,3.8) -- (0.8,3) -- (0,3);
  \node[anchor=west] at (0.9,3.4) {GT A};
  \filldraw[fill=cyan!50!blue!50] (0,2) -- (0,2.8) -- (0.8,2.8) -- (0.8,2) -- (0,2);
  \node[anchor=west] at (0.9,2.4) {GT B};
  \draw[pattern=north west lines, pattern color=black!50] (0,1) -- (0,1.8) -- (0.8,1.8) -- (0.8,1) -- (0,1);
  \node[anchor=west] at (0.9,1.4) {PD A};
  \draw (0,0) -- (0,0.8) -- (0.8,0.8) -- (0.8,0) -- (0,0);
  \node[anchor=west] at (0.9,0.4) {PD B};
\end{tikzpicture}
\begin{tikzpicture}[scale=0.7]
	\coordinate (A) at (0,4);
  \coordinate (B) at (0,0);
  \coordinate (C) at (4,0);
  \coordinate (D) at (4,4);
  \coordinate (E) at (2,4);
  \coordinate (F) at (2,0);
  \filldraw[fill=cyan!50!blue!50] (A) -- (B) -- (C) -- (D);
  \filldraw[fill=white!30!yellow] (A) -- (E) -- (4,3) -- (0,1) -- (F) -- (B) -- (A);
  \draw[pattern=north west lines, pattern color=black!50, dashed] (A) -- (E) -- (F) -- (B) -- (A);
  \draw[semithick] (A) -- (B);
  \draw[semithick] (B) -- (C);
  \draw[semithick] (C) -- (D);
  \draw[semithick] (D) -- (A);
\end{tikzpicture}
\begin{tikzpicture}[scale=0.7]
	\coordinate (A) at (0,4);
  \coordinate (B) at (0,0);
  \coordinate (C) at (4,0);
  \coordinate (D) at (4,4);
  \coordinate (E) at (2,4);
  \coordinate (F) at (2,0);
  
  \filldraw[fill=cyan!50!blue!50] (A) -- (B) -- (C) -- (D);
  \filldraw[fill=white!30!yellow] (A) -- (E) -- (4,3.5) -- (0,2.5) -- (4,1.5) -- (0,0.5) -- (F) -- (B) -- (A);
  \draw[pattern=north west lines, pattern color=black!50, dashed] (A) -- (E) -- (F) -- (B) -- (A);
  \draw[semithick] (A) -- (B);
  \draw[semithick] (B) -- (C);
  \draw[semithick] (C) -- (D);
  \draw[semithick] (D) -- (A);
\end{tikzpicture}
\begin{tikzpicture}[scale=0.7]
	\coordinate (A) at (0,4);
  \coordinate (B) at (0,0);
  \coordinate (C) at (4,0);
  \coordinate (D) at (4,4);
  \coordinate (E) at (2,4);
  \coordinate (F) at (2,0);
  \filldraw[fill=cyan!50!blue!50] (A) -- (B) -- (C) -- (D);
  \filldraw[fill=white!30!yellow] (A) -- (E) -- (4,3.75) -- (0,3.25) -- (4,2.75) -- (0,2.25) -- (4,1.75) -- (0,1.25) -- (4,0.75) -- (0,0.25) -- (F) -- (B) -- (A);
  \draw[pattern=north west lines, pattern color=black!50, dashed] (A) -- (E) -- (F) -- (B) -- (A);
  \draw[semithick] (A) -- (B);
  \draw[semithick] (B) -- (C);
  \draw[semithick] (C) -- (D);
  \draw[semithick] (D) -- (A);
\end{tikzpicture}
\begin{tikzpicture}
    \begin{axis}[
        width=0.38\textwidth,
        height=0.25\textwidth,
        xticklabels={},
        ylabel={score (\%)},
        ymax=61,
        ylabel near ticks, ylabel shift={-5pt},
        legend columns=2,
        legend style={at={(0.41,-0.05)},anchor=north}
    ]
    \addplot table [x, y, col sep=comma] {data/pnmiou.csv};
    \addplot [domain=1:5,dashed] {60};
    \legend{{PNM IoU}, {mIoU}}
    \end{axis}
\end{tikzpicture}
\caption{
Example showing mean IoU is insensitive to sharp boundaries.
Left: the same prediction with different ground truth leads to the same mean IoU score on different images.
``PD A'' stands for ``prediction of class A" and ``GT A'' stands for ``ground truth of class A''.
Please refer to the text for details.
Right: PNM and mean IoU scores of such image series.
As the boundary gets sharper, the mean IoU stays the same showing its insensitivity.
In contrast, PNM IoU decreases rapidly.
Best viewed in color.
}
\label{fig:pnm_iou}
\end{figure}

We further point out a sharp boundary identifiability problem for the popular metric, \emph{mean intersection over union (mIoU)}, and metrics based on IoU scores like \emph{panoptic quality (PQ)}~\cite{panoptic_seg}.
Specifically, consider a series of square images constructed by the following process.
For the $n$-th image, starting from the midpoint of the upper edge, connect the first point of $2^{n+1}$-section of the right edge, the third point of $2^{n+1}$-section of the left edge, ..., the midpoint of the bottom edge.
Let the semantic label of the left part be A and B for the right.
\figurename~\ref{fig:pnm_iou} shows the first three images of the series.

For this series of images, we assume a trivial segmenter that simply horizontally divides an image into equal parts and predicts A for the left part and B for the right.
Then we can calculate mIoU for the trivial segmenter as 
\begin{equation}
\label{eq:mIoU}
\text{mIoU} = \frac{1}{K}\sum_{k=1}^K\frac{|\text{PD}_k\cap \text{GT}_k|}{|\text{PD}_k\cup \text{GT}_k|} = \frac{1}{K}\sum_{k=1}^K \frac{\sum_{i}\delta(y_i, t_i)\delta(t_i, k)}{\sum_{i}\min(\delta(y_i, k)+\delta(t_i, k), 1)},
\end{equation}
where $k\in\{1, \dots, K\}$ is the index of the semantic classes, ``GT'' is the ground truth segmentation mask, ``PD'' is the predicted mask, $\delta(\cdot, \cdot)$ is the Kronecker delta defined in Eq.~\eqref{eq:pnm}, $y_i$ is the prediction, and $t_i$ is the label of pixel $i$ where $y_i, t_i\in \{1, \dots, K\}$.
Eq.~\eqref{eq:mIoU} indicates the mIoU of the classes is $60\%$ for the entire image series, as the area of correctly predicted pixels is the same for all images.

To solve the problems above, we propose a new evaluation metric, \textit{PNM IoU} based on the PNM pixel weights.
In specific, we weight the calculation of IoU based on PNM as
\begin{equation}
\text{PNM IoU}(d) = \frac{1}{K}\sum_{k=1}^K\frac{\sum_{i}\delta(y_i, t_i)\delta(t_i, k)\cdot\rho_i(d)}{\sum_{i}\min(\delta(y_i, k)+\delta(t_i, k), 1)\cdot\rho_i(d)}.
\end{equation}
We empirically calculate the PNM IoU for the trivial segmenter of the first five images of the series, and \figurename~\ref{fig:pnm_iou} shows the results. 
Different from mIoU, PNM IoU rapidly decreases as the image gets ``sharper'' showing that PNM IoU may offer better indication of the segmenter's performance, especially on error-prone boundary pixels.

\section{Experiments}
\label{sec:exp}

We report our experimental results on multiple datasets of three segmentation tasks.
We first briefly describe our experimental settings and datasets.
Next, we study the effects of hyperparameters of PNM.
Finally, we compare PNM with SOTA methods on semantic, instance, and panoptic segmentation\footnote{Due to space limits, codes, running logs, and additional experimental results such as qualitative examples and speed analysis are attached in the supplementary material.}.

\subsection{Experiment Settings and Datasets}
\label{sec:settings_exp}

\subsubsection{Settings.}

We conduct the experiments on Nvidia A100 GPUs with 40GB memories so that all baseline models can fit into memory.
Note that PNM will not increase the memory consumption as we can compute the PNM pixel weights in advance.
Unless otherwise stated, we train the networks on 4 GPUs in parallel.
For the semantic segmentation tasks, we train the models for 160k batches at most.
For instance and panoptic segmentation tasks, we train the models for 12 epochs.
In all experiments, we fix our random seed to 0 to make our results reproducible, and we do not use test-time augmentation to exclude the effects from other factors unless for a fair comparison with baselines.

\subsubsection{Datasets.}

For semantic segmentation, we conduct experiments on both Cityscapes~\cite{cityscapes} and ADE20K~\cite{ade20k} datasets.
For instance and panoptic segmentation, we use MS COCO~\cite{mscoco} dataset.
Cityscapes is a street view dataset with 19 classes and around 5,000 fine annotated images.
For most models, we first perform data augmentations on the input images and then crop them to a size of 512$\times$1024 for training with a batch size of 2.
ADE20K comprises 150 classes and more than 27K images from the SUN and Places databases.
We crop ADE20K images to 512$\times$512 and set the batch size to 4.
MS COCO is a large-scale segmentation and detection benchmark dataset for common objects, with 80 classes for instance segmentation, 80 thing classes and 53 stuff classes for panoptic segmentation.
For all models on this dataset, we set the batch size to 2.

\subsection{Effects of Hyperparameters}
\label{sec:ablation_exp}

\begin{figure}[!tb]
\centering
\begin{tikzpicture}
\begin{groupplot}[
    group style={
        group size=1 by 2,
        xticklabels at=edge bottom,
        vertical sep=0pt,
    },
]
\nextgroupplot[
    ymin=79.6,
    axis x line=top,
    axis y discontinuity=parallel,
    height=0.3\textwidth,
    width=0.5\textwidth,
    ytick={79.8,80,80.2,80.4},
    xtick={1,2,...,11},
    title={(a) Effect of locality scale $d$.},
    title style={yshift=-5pt},
    xmin=0,xmax=12,
]
\addplot table [x, y, col sep=comma] {data/ablation_scale.csv};
\nextgroupplot[
    ymin=77,
    ymax=77.2,
    ytick={77.05, 77.2, 77.12},
    axis x line=bottom,
    height=0.2\textwidth,
    width=0.5\textwidth,
    xtick={1,2,...,11},
    xticklabels={3, 11, 21, 31, 41, 51, 61, 71, 81, 91, 101},
    xmin=0,xmax=12,
    xlabel={Locality scale $d$},
    xticklabel style={rotate=45},
    ylabel={Mean IoU (\%)},
    ylabel near ticks, ylabel shift={-5pt},
    legend style={at={(0.82,1.9)}},
    every axis y label/.append style={at=(ticklabel cs:1.5)}],
]
\addplot table [x, y, col sep=comma] {data/ablation_scale.csv};
\legend{\footnotesize{PNM}, \footnotesize{DeepLabV3}}
\addplot [domain=1:11,dashed] {77.12};
\end{groupplot}
\end{tikzpicture}
\begin{tikzpicture}
    \begin{axis}[
        width  = 0.5\textwidth,
        height = 0.405\textwidth,
        major x tick style = transparent,
        ybar=\pgflinewidth,
        bar width=8pt,
        ymin = 78,
        ylabel = {Mean IoU (\%)},
        ylabel near ticks, ylabel shift={-5pt},
        symbolic x coords={CCNet,PSPNet,DeepLabV3},
        xtick = data,
        enlarge x limits=0.25,
        legend columns=4,
        title={(b) Effect of the transformations.},
        title style={yshift=-5pt},
        legend style={at={(0.85,0.2)}}
    ]
        \addplot+[style={bblue,fill=bblue,mark=none}]
            table [x, y, col sep=comma] {data/ablation_log.csv};

        \addplot+[style={rred,fill=rred,mark=none}]
            table [x, y, col sep=comma] {data/ablation_inverse.csv};

        \addplot+[style={ggreen,fill=ggreen,mark=none}]
            table [x, y, col sep=comma] {data/ablation_completion.csv};

        \legend{$\rho^{(1)}$, $\rho^{(2)}$, $\rho^{(3)}$}
\end{axis}
\end{tikzpicture}

\caption{Hyperparameter study of PNM on Cityscapes \textit{val} with 40k training iterations.
}
\label{fig:ablation}
\end{figure}
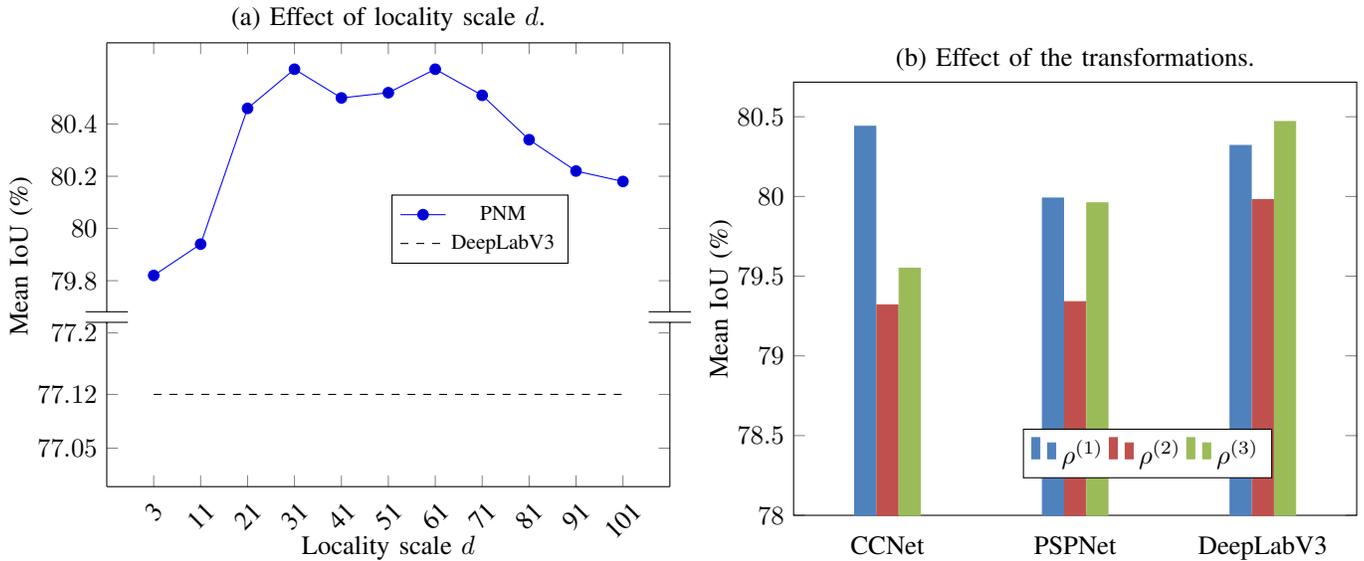

We study the effect of different locality scale and the transformation method (i.e., Eq.~\eqref{eq:pnm_weight_1}, \eqref{eq:pnm_weight_2}, and \eqref{eq:pnm_weight_3}) from PNM to pixel weight.
We train and evaluate on Cityscapes \textit{val}, and \figurename~\ref{fig:ablation} shows the result.

\figurename~\ref{fig:ablation}(a) shows the mean IoU of different locality scale $d$.
We find that by adding PNM to DeepLabV3, the mean IoU is significantly improved compared to the baseline (DeepLabV3) for a wide range of choices of $d$.
Meanwhile, we observe that as $d$ increases beyond 61, the mIoU goes down.
The reason is that too large $d$ (especially when it is comparable to the image size) defeats the PNM's purpose of capturing only local information.

\figurename~\ref{fig:ablation}(b) shows the impact of different transformations on PNM.
We find that the scheme in Eq.~\eqref{eq:pnm_weight_1} assigns pixel weights more mildly and is more stable than the other two.
Based on these observations, we use $d=35$ and Eq.~\eqref{eq:pnm_weight_1} as the transformation in all following experiments.

\subsection{Semantic Segmentation}
\label{sec:semantic_seg_exp}

\subsubsection{Comparison with SOTA Segmenters.}

We study the performance of PNM with different training durations, model sizes, and network architectures on ADE20K and Cityscapes.
The comparison baselines include FCN~\cite{fcn}, PSPNet~\cite{pspnet}, DeepLabV3~\cite{deeplabv3}, NLNet~\cite{nlnet}, CCNet~\cite{ccnet}, OCRNet~\cite{ocrnet}, and Segformers~\cite{segformer}.
Table~\ref{tb:ade20k_val_total} and \ref{tb:cityscapes_val_total} summarize the effect on models of different sizes on ADE20K, and report the effect of PNM on the basis of different semantic methods with different training durations on Cityscapes.

\begin{table}[!tb]
\centering
\begin{tabular}{@{}lccccc@{}}
\toprule
\multirow{2}{*}{Method} & \multirow{2}{*}{Backbone} & \multicolumn{2}{c}{Mean IoU (\%)} & \multicolumn{2}{c}{PNM IoU (\%)} \\ \cmidrule(l){3-6} 
                        &                           & w/o PNM        & with PNM                & w/o PNM        & with PNM        \\ \midrule
CCNet                   & ResNet-50                 & 42.08          & \textbf{43.40 (+1.32)}  & 33.91          & \textbf{35.08 (+1.17)}  \\
CCNet                   & ResNet-101                & 43.71          & \textbf{44.84 (+1.13)}  & 35.14          & \textbf{36.11 (+0.97)}  \\
DeepLabv3               & ResNet-50                 & \textbf{44.08} & 43.35 (-0.73)           & 34.40          & \textbf{35.04 (+0.64)}  \\
DeepLabv3               & ResNet-101                & 45.00          & \textbf{46.17 (+1.17)}  & 36.21          & \textbf{37.23 (+1.02)}  \\
PSPNet                  & ResNet-50                 & 42.48          & \textbf{43.25 (+0.77)}  & 34.37          & \textbf{34.95 (+0.58)}  \\
PSPNet                  & ResNet-101                & 44.39          & \textbf{45.23 (+0.84)}  & 35.79          & \textbf{36.53 (+0.74)}  \\
OCRNet                  & HRNet-18                  & 39.32          & \textbf{41.38 (+1.06)}  & 32.02          & \textbf{33.70 (+1.68)}  \\
OCRNet                  & HRNet-48                  & 43.25          & \textbf{44.88 (+1.63)}  & 35.01          & \textbf{36.37 (+1.36)}  \\
Segformer*               & MiT-B2                    & 45.58          & \textbf{46.25 (+0.67)} & 36.77          & \textbf{37.40 (+0.63)}  \\
Segformer*               & MiT-B5                    & 49.13          & \textbf{49.56 (+0.46)} & 39.44          & \textbf{39.73 (+0.29)}  \\ \bottomrule
\end{tabular}
\caption{Comparison results on ADE20k \textit{val}.
We do not use test-time augmentations.
* We train the Segformers with PNM on 4 GPUs and obtain a better result than original Segformers trained on 8 GPUs ($2\times$ larger batch size).}
\label{tb:ade20k_val_total}
\end{table}

\begin{table}[!tb]
\centering
\begin{tabular}{@{}lcccccc@{}}
\toprule
\multirow{2}{*}{Method} & \multirow{2}{*}{Backbone} & \multirow{2}{*}{\#Trains} & \multicolumn{2}{c}{Mean IoU (\%)}       & \multicolumn{2}{c}{PNM IoU (\%)}         \\ \cmidrule(l){4-7} 
                        &                           &                           & w/o PNM        & with PNM               & w/o PNM         & with PNM               \\ \midrule
FCN                     & ResNet-101                & 40k                       & 75.45          & \textbf{75.65 (+0.20)} & 59.65           & \textbf{60.22 (+0.57)} \\
FCN                     & ResNet-101                & 160k                  & 75.13          & \textbf{79.08 (+3.95)} & 59.57           & \textbf{63.04 (+3.47)} \\
CCNet                   & ResNet-101                & 40k                       & 76.35          & \textbf{80.44 (+4.09)} & 60.52           & \textbf{64.18 (+3.66)} \\
CCNet                   & ResNet-101                & 160k                  & 78.87          & \textbf{80.40 (+1.53)} & 62.64           & \textbf{63.04 (+0.40)} \\
NLNet                   & ResNet-101                & 40k                       & 78.66          & \textbf{79.37 (+0.71)} & 62.44           & \textbf{63.34 (+0.90)} \\
NLNet                   & ResNet-101                & 160k                      & 78.93          & \textbf{79.83 (+0.90)} & 62.82           & \textbf{63.74 (+0.92)} \\
PSPNet                  & ResNet-101                & 40k                       & 78.34          & \textbf{79.99 (+1.65)} & 62.37           & \textbf{64.00 (+1.63)} \\
PSPNet                  & ResNet-101                & 160k                  & 79.76          & \textbf{80.05 (+0.29)} & 63.56           & \textbf{63.99 (+0.43)} \\
DeepLabv3               & ResNet-101                & 40k                       & 77.12          & \textbf{80.32 (+3.20)} & 62.37           & \textbf{64.09 (+1.72)} \\
DeepLabv3               & ResNet-101                & 160k                  & 80.20          & \textbf{80.47 (+0.27)} & 63.83           & \textbf{64.18 (+0.35)} \\
OCRNet                  & HRNet-48                  & 40k                       & 80.58          & \textbf{81.34 (+0.77)} & 64.03           & \textbf{64.89 (+0.86)} \\
OCRNet                  & HRNet-48                  & 160k                      & 81.35          & \textbf{81.57 (+0.22)} & 64.75           & \textbf{65.12 (+0.37)} \\
Segformer*              & MiT-B2                    & 160k                      & \textbf{81.08} & \textbf{81.08}         & 64.55           & \textbf{64.76 (+0.21)} \\
Segformer*              & MiT-B5                    & 160k                      & \textbf{82.25} & 82.08 (-0.17)          & 65.38           & \textbf{65.47 (+0.09)} \\ \bottomrule
\end{tabular}
\caption{Comparison results on Cityscapes \textit{val}.
We do not use extra training data or test-time augmentation.
* For segformers, we train on 8 GPUs and crop images to $1024\times 1024$ patches
($\sim$4 times larger input than other models) for fair comparison.
}
\label{tb:cityscapes_val_total}
\end{table}

As \tablename~\ref{tb:ade20k_val_total} shows, PNM significantly improves the mean IoU score of almost all models.
Meanwhile, PNM significantly improves models of different sizes, indicating that both PNM and larger model capacity can improve the segmentation performance without conflicting with each other, and PNM taps more potentials of the model.
PNM also brings improvements in various network architectures, including both transformers and CNN backbones, as well as different segmentation heads.
Thus, we can see that PNM pixel weights are essential and generally applicable, independent of the model choice. 
\tablename~\ref{tb:cityscapes_val_total} also confirms that PNM also improves mIoU of almost all models on Cityscapes, indicating the generality of PNM on different semantic segmentation datasets.

We also try to decrease the training time to 1/4 (40K from 160K).  
\tablename~\ref{tb:cityscapes_val_total} also shows that PNM can exceed the performance of original models, even with 1/4 training time.  Intuitively, PNM can simplify the learning process by assigning larger weights to error-prone regions.

\tablename~\ref{tb:cityscapes_val_total} shows little improvement of mIoU for Segformers.
By empirically analyzing the predicted images, we find the reason is PNM encourages the model to make finer segmentation, which may lead to a prediction of unseen classes in the ground truth and in turn greatly reduces the mIoU score\footnote{We give qualitative examples of the Segformers in the supplements.}.
We suspect the model is approaching the performance ceiling on Cityscapes, as on the larger ADE20K dataset, PNM significantly improves the mIoU of Segformers.  

For PNM IoU, all models incorporating PNM, including Segformers, show significant improvements on Cityscapes. 
This result indicates that various segmenters of different network architectures can benefit from PNM by making finer segmentation in error-prone regions.



\subsubsection{Comparison with Boundary-based Segmenters.}

\begin{table}[!tb]
\parbox{.5\linewidth}{
\centering
\begin{tabular}{@{}lccc@{}}
\toprule
Model                                                    & ms+flip & Backbone   & Mean IoU         \\ \midrule
Baseline                                                 & \cmark  & ResNet-101 & 79.6          \\
GC-SCNN                                                  & \cmark  & ResNet-101 & 81.0          \\
\begin{tabular}[c]{@{}l@{}}GC-SCNN\\ +Segfix\end{tabular}& \cmark  & ResNet-101 & 81.5          \\
\midrule
PNM                                                      & \cmark  & ResNet-101 & \textbf{81.6} \\ \bottomrule
\end{tabular}
}
\hfill
\parbox{.46\linewidth}{
\centering
\begin{tabular}{@{}lccc@{}}
\toprule
Model                                                      & ms+flip  & Backbone   & Mean IoU          \\ \midrule
Baseline                                                   & \xmark & ResNet-101 & 44.8          \\
\begin{tabular}[c]{@{}l@{}}Baseline\\ +Segfix\end{tabular} & \xmark & ResNet-101 & 45.4          \\
\midrule
PNM                                                        & \xmark & ResNet-101 & \textbf{46.2} \\ 
PNM                                                     & \cmark & ResNet-101 & \textbf{47.9} \\ \bottomrule
\end{tabular}
}
\caption{
Comparison with boundary-based methods.
Baseline is DeepLabV3.
``ms+flip'' means multiscale and random flip augmentation for test.
Left: comparison results on Cityscapes \textit{val}.
Right: comparison results on ADE20K \textit{val}.
}
\label{tb:boundary}
\end{table}

We compare PNM with strong boundary-aware segmenters GC-SCNN~\cite{DBLP:conf/iccv/TakikawaAJF19} and Segfix~\cite{segfix} on both ADE20K and Cityscapes.
\tablename~\ref{tb:boundary} shows the results.

PNM outperforms both boundary-based methods on both datasets.
Specifically, the left shows that PNM leads to better segmentation results than both Segfix and GC-SCNN, and even outperforms the combination of the two methods.
The right indicates that the performance margin even increases when comparing PNM with the baselines on a larger dataset.
It also shows that the PNM-based model can also greatly benefit from test-time augmentations.
Given that boundary-based methods require non-trivial additional computational cost, we argue PNM is more effective and efficient for segmentation refinement.

\subsection{Instance and Panoptic Segmentation}
\label{sec:instance_seg_exp}

We conduct experiments on instance and panoptic segmentation on MS COCO dataset.
For instance segmentation, we adopt mask R-CNN~\cite{mask-rcnn} as the baseline and apply PNM to its segmentation branch.
For panoptic segmentation, we use panoptic FPN~\cite{panoptic_fpn} as the baseline.
\tablename~\ref{tb:instance_seg} and~\ref{tb:panoptic_seg} show the comparison.

PNM improves the performance of baselines on both instance and panoptic segmentation.
The improvement of instance segmentation is less significant, and we believe two reasons are reducing the impact of pixel weights: 1) the size of the regions of interest (RoIs) is so small (28$\times$28) that we can only use a very small $d=5$, and 2) the model only predicts binary masks for each pixel, which reduces the variance of PNM.
For panoptic segmentation, PNM significantly increases the panoptic quality (PQ) on both things and stuff.
Meanwhile, although the original intention of the PNM design is to improve the segmentation quality (SQ), the recognition quality (RQ) can also benefit from PNM.

In summary, the superior effect of PNM on three datasets indicates that PNM can consistently improve the model segmentation performance for various input images in all task settings and can be plugged into most existing models without adding heavy computational overhead.

\begin{table}[!tb]
\parbox{.4\linewidth}{
\centering
\begin{tabular}{@{}lccc@{}}
\toprule
Method   & mAP           & mAP50         & mAP75         \\ \midrule
Baseline & 35.6          & \textbf{56.6} & 38.2          \\
PNM      & \textbf{35.7} & \textbf{56.6} & \textbf{38.3} \\ \bottomrule
\end{tabular}
\caption{Instance segmentation results.
Baseline is mask R-CNN with Resnet-101 backbone.}
\label{tb:instance_seg}
}
\hfill
\parbox{.56\linewidth}{
\centering
\begin{tabular}{@{}lccccc@{}}
\toprule
\multirow{2}{*}{Method} & \multirow{2}{*}{RQ(\%)} & \multirow{2}{*}{SQ(\%)} & \multicolumn{3}{c}{PQ (\%)}                      \\ \cmidrule(l){4-6} 
                        &                         &                         & things         & stuff          & total          \\ \midrule
Baseline                & 49.78                   & 78.12                   & 48.47          & 29.22          & 40.80          \\
PNM                     & \textbf{50.44}          & \textbf{78.27}          & \textbf{48.83} & \textbf{30.11} & \textbf{41.37} \\ \bottomrule
\end{tabular}
\caption{
Panoptic segmentation results.
Baseline is panoptic FPN with Resnet-101 backbone.}
\label{tb:panoptic_seg}
}
\end{table}

\section{Conclusion and Future Work}
\label{sec:conclusion}

Taking a closer look at image segmentation, despite the overall performance, boundary and small object prediction remain challenges even for SOTA models. Even worse, the current evaluation metric, mIoU does poorly in 
reflecting actual boundary classification accuracy.  
Inspired by methods in community detection, this paper presents the pixel null model, a pixel-level prior distribution of correct segmentation probability that helps the segmenters to focus on error-prone regions.
We show that PNM correctly captures the misclassification distribution of SOTA segmenters.
We also show that PNM-based metric overcomes the sharp boundary identifiability issue of IoU-based metrics.
The significant improvement over SOTA segmentation networks on three image segmentation tasks and three benchmark datasets demonstrate the general effectiveness of PNM.

For future work, we would like to explore the application of PNM to tasks such as video segmentation, keypoint detection, and 3D reconstruction. Also, PNM is currently calculated on the ground truth segmentation mask, which requires heavy manual labeling of pixels.
It is worth exploring how to generate high-quality PNM with simple unsupervised models.
In this way, we can apply PNM to the evaluation process or the vast unlabeled data in the wild.

%
%
\bibliographystyle{plain}
\bibliography{arxiv_pnm}

\end{document}